\pdfoutput=1

\documentclass[11pt]{article}

\usepackage{acl}

\usepackage{times}
\usepackage{latexsym}

\usepackage[T1]{fontenc}

\usepackage[utf8]{inputenc}

\usepackage{microtype}

\usepackage{inconsolata}

\usepackage{graphicx}
\usepackage{booktabs}
\usepackage{multirow}
\usepackage{amsmath}
\usepackage{amssymb}
\usepackage{pifont}
\usepackage{multirow}
\usepackage{algorithm}
\usepackage{algpseudocode}
\usepackage{enumitem}
\usepackage{CJKutf8}

\newcommand{\ja}[1]{\begin{CJK}{UTF8}{ipxm}#1\end{CJK}}

\newcommand{\xmark}{\text{\ding{55}}}
\newcommand{\uline}[1]{\underline{#1}}

\title{CADEL: A Corpus of Administrative Web Documents\\for Japanese Entity Linking}

\author{
Shohei Higashiyama
\hspace{0.1cm}
Masao Ideuchi
\hspace{0.1cm}
Masao Utiyama\\
National Institute of Information and Communications Technology\\
\texttt{\{shohei.higashiyama,masao.ideuchi,mutiyama\}nict.go.jp}
}

\begin{document}
\maketitle
\begin{abstract}
Entity linking is the task of associating linguistic expressions with entries in a knowledge base that represent real-world entities and concepts.
Language resources for this task have primarily been developed for English, and the resources available for evaluating Japanese systems remain limited.
In this study, we develop a corpus design policy for the entity linking task and construct an annotated corpus for training and evaluating Japanese entity linking systems, with rich coverage of linguistic expressions referring to entities that are specific to Japan.
Evaluation of inter-annotator agreement confirms the high consistency of the annotations in the corpus, and a preliminary experiment on entity disambiguation based on string matching suggests that the corpus contains a substantial number of non-trivial cases, supporting its potential usefulness as an evaluation benchmark.\footnote{
This paper is based on an English translation of \cite{cadel-ja}, originally published in IPSJ SIG Technical Report, Vol.~2024-NL-260, No.~10, pp.~1--15, 2024, with its organization revised and some content added and modified, mainly in \S\ref{sec:discussion}. The copyright of the original Japanese paper is held by the Information Processing Society of Japan.}
\end{abstract}

\section{Introduction}
Real-world entities, including specific objects and concepts, are referred to by linguistic expressions in text.
Entity linking (EL)~\cite{shen-2014-entity,martinez-2020-information} addresses the problem of linking such expressions to the entities they refer to.
It has been studied in natural language processing and the Semantic Web.

EL can be divided into two subproblems: mention recognition 
and entity disambiguation.
Mention recognition is the process of identifying linguistic expressions in text that refer to some entity, and in many cases, also identifying the entity type or category.
A linguistic expression that refers to an entity is called a \textit{mention}; when the expression is primarily a proper name, it is also called a \textit{named entity} (NE).
Entity disambiguation is the process of mapping mentions in text to appropriate entries in a knowledge base (KB) that contains entries corresponding to various real-world entities, typically Wikipedia~\cite{hachey-2013-evaluating} or Wikidata~\cite{moller-2022-survey}.
The EL setting that uses Wikipedia as the KB is in particular called Wikification~\cite{mihalcea-2007-wikify}.

EL has a wide range of applications, including question answering that exploits disambiguation results for input text~\cite{dimitrakis-2020-survey}, KB completion using disambiguated entity knowledge~\cite{ji-grishman-2011-knowledge}, and representation learning~\cite{ri-etal-2022-mluke} and language generation~\cite{yu-2022-survey} based on acquired knowledge or constructed KBs.

Since the late 2000s, research has been conducted primarily for English on the construction of language resources for EL~\cite{cucerzan-2007-large,mcnamee-etal-2010-evaluation,hoffart-etal-2011-robust,ratinov-etal-2011-local,hoffart-2012-kore,roder-etal-2014-n3,rizzo-etal-2016-making,rosales-2018-voxel,provatorova-etal-2021-robustness}, as well as comparative studies across benchmarks for system evaluation using those resources~\cite{ling-etal-2015-design,van-erp-etal-2016-evaluating,roder-etal-2018-gerbil,rosales-mendez-etal-2019-fine,milich-akbik-2023-zelda,bast-etal-2023-fair}.
In contrast, language resources constructed for Japanese EL remain limited~\cite{jargalsaikhan-etal-2016-building,murawaki-mori-2016-wikification,sekine-2023}, and it is still rare to compare systems using a shared benchmark.

Recent advances in deep learning and foundation models have enabled large-scale models trained primarily on English-centric data to demonstrate strong multilingual processing capabilities.
Nevertheless, for the Japanese EL task, it remains important to develop systems that can appropriately analyze linguistic expressions and link them to Japan-specific concepts.
For the training and evaluation of such systems, it is therefore desirable to have access to a Japanese corpus rich in Japanese-language expressions.

\begin{table*}[t]
\centering
\small
\begin{tabular}{lcccrc} \toprule
Dataset & Lang. & Data source & KB & \multicolumn{1}{c}{\#Mentions} & Manual \\
\midrule
CADEL (Ours) & ja & Web & Wikidata & 8,082 & $\checkmark$ \\
jawikify~\cite{jargalsaikhan-etal-2016-building} & ja & News & Wikipedia & 25,675 & $\checkmark$ \\
JWC~\cite{murawaki-mori-2016-wikification} & ja & Blog, Twitter, etc. & Wikipedia & 8,057 & $\checkmark$ \\
Shinra 2023 dataset~\cite{sekine-2023} & ja & Wikipedia & Wikipedia & 59,429 & $\triangle$ \\
\midrule
WikiANN~\cite{pan-etal-2017-cross} & multi & Wikipedia & Wikipedia & 1.9M & $\xmark$ \\
Mewsli-9~\cite{botha-etal-2020-entity} & multi & Wikinews & Wikidata & 34,463 & $\xmark$ \\
DaMuEL~\cite{kubesa-2023-damuel} & multi & Wikipedia & Wikidata & 54.9M & $\xmark$ \\
\bottomrule
\end{tabular}
\caption{Characteristics of EL datasets with Japanese text. For multilingual (multi) datasets, we report the number of mentions only for the Japanese portion. In the Manual Column, ``$\checkmark$'', ``$\triangle$'', and ``$\xmark$'' indicate full, partial, and no manual correction, respectively.}
\label{tab:datasets}
\end{table*}

To this end, we constructed CADEL, an annotated corpus for Japanese EL derived from administrative web documents.
In constructing this corpus, we developed a design policy based on key issues in defining the EL task, and clarified how to handle cases in which it is difficult to determine whether a mention should be linked to an entry (``related link'' in \S\ref{sec:policy_entry}).
The resulting corpus consists of 160 articles, mainly public relations magazines and white papers issued by Japanese ministries and agencies, and contains 6,939 named mentions and 1,143 non-named mentions, together with coreference relations among them and linked Wikidata entry IDs.

On a subset of 10 articles from the corpus, we measured inter-annotator agreement and found generally high agreement rates (mentions: F1 score of at least 0.79; coreference: LEA~\cite{moosavi-strube-2016-coreference} of at least 0.91; linking: F1 score of at least 0.83 for ``exact link'' cases), confirming that the corpus annotations are highly consistent.

We also conducted a preliminary disambiguation experiment on all named mentions in the corpus using exact string matching and heuristics.
The results suggest that, while many cases can be resolved by such simple methods, a substantial number of cases (about 1,240 mentions) remains non-trivial.
Based on this observation, we created an evaluation-oriented data split that priorities such difficult cases, which we expect to be useful for future research on Japanese EL.
The corpus is publicly available at \url{https://github.com/shigashiyama/cadel}.

\section{Related Work} \label{sec:rel_work}
\subsection{Datasets for Japanese EL}
Several studies have constructed language resources for Japanese EL~\cite{jargalsaikhan-etal-2016-building,murawaki-mori-2016-wikification,sekine-2023} and those for multilingual EL including Japanese~\cite{pan-etal-2017-cross,botha-etal-2020-entity,kubesa-2023-damuel}.
Table~\ref{tab:datasets} summarizes the characteristics of these existing datasets and CADEL.

Jawikify\footnote{\url{http://www.cl.ecei.tohoku.ac.jp/jawikify/}} by~\citet{jargalsaikhan-etal-2016-building} and the Japanese Wikification Corpora (JWC)\footnote{\url{https://murawaki.org/research/wikify-data.html} (as of March 2026, the corpus is not publicly available).} by~\citet{murawaki-mori-2016-wikification} are both manually annotated corpora constructed for the Japanese Wikification task.
\citet{jargalsaikhan-etal-2016-building} used the newspaper text derived from the BCCWJ\footnote{\url{https://clrd.ninjal.ac.jp/bccwj/en/}} included in the Extended Named Entity-Tagged Corpus,\footnote{\url{https://www.gsk.or.jp/catalog/gsk2014-a/}} and manually assigned Wikipedia entry (i.e., article) information to pre-annotated mentions.
\citet{murawaki-mori-2016-wikification} manually annotated mentions and Wikipedia entry information for white paper and Yahoo! Blog data in the BCCWJ, as well as Twitter (now X) data.
Murawaki et al.\ adopt a policy of targeting entities of any topic, as long as a corresponding entry exists.

\citet{sekine-2023,sekine-2024} organized a series of shared tasks on \textit{linking attribute values} in Japanese Wikipedia articles within the Shinra project.\footnote{\url{http://shinra-project.info/}}
Under attributes (e.g., ``work'') defined for each category (e.g., ``person name'') in the Extended Named Entity (ENE) hierarchy~\cite{sekine-etal-2002-extended}, linking attribute values refers to the task of selecting an appropriate Wikipedia article (e.g., the article ``Arashi\_(novel)'') for an attribute-value mention (e.g., ``Arashi,'' of the category ``work'') appearing in a Wikipedia article (e.g., the article on ``Shimazaki Toson'' with the category ``person name'').
Manual annotation data are used for evaluation in this task.\footnote{We inquired with the organizers of the Shinra project and received confirmation that manually created test data are used for evaluation. The data are publicly available at \url{http://shinra-project.info/shinra/data/}, although the gold labels for the test set are not released.}

\citet{pan-etal-2017-cross} constructed WikiANN,\footnote{\url{https://elisa-ie.github.io/wikiann/}} a cross-lingual EL dataset in which mentions in Wikipedia articles in 282 languages are linked to English Wikipedia articles.
The mentions and cross-lingual links were assigned automatically by performing named entity recognition on source-language articles, filtering NE-annotated article snippets, and aligning article-title pairs between each source language and the target language (English).

\citet{botha-etal-2020-entity} proposed a multilingual EL task in which mentions in a multilingual collection of texts are linked to entries in a language-independent KB, and constructed Mewsli-9,\footnote{\url{https://github.com/google-research/google-research/tree/master/dense_representations_for_entity_retrieval/mel}} a multilingual dataset consisting of Wikinews articles in nine languages.
Anchor texts, to which Wikipedia links had been added by article editors, were treated as mentions, and Wikidata entry\footnote{In Wikidata, all content objects are called ``entities,'' while those corresponding to real-world objects are called ``items.'' In this paper, we refer to Wikidata items as ``entries.''} IDs were assigned on the basis of interwiki links (links across Wikimedia sites).

\citet{kubesa-2023-damuel} constructed DaMuEL,\footnote{\url{https://hdl.handle.net/11234/1-5047}} a dataset in which Wikidata entry information is assigned to Wikipedia articles in 53 languages on the basis of links between Wikipedia articles and interwiki links.
While Wikipedia's editorial style is to place article links only on the first mention of each concept (entity) referred to in the body text, they sought to improve coverage by automatically assigning links to subsequent mentions as well, when they met heuristic criteria.

\paragraph*{Positioning of this study}
Although automatically constructed multilingual datasets are useful as large-scale silver-standard data, they are not necessarily accompanied by quality evaluation, and concerns remain regarding annotation consistency and coverage.
Compared with Jawikify, which is also a manually annotated corpus like ours, CADEL is characterized by the following points:
(1) it was constructed on the basis of an explicit and comprehensive discussion of design considerations, including the distinction between exact and related links in \S\ref{sec:policy_entry};
(2) the publication years of the source texts are more recent (Jawikify: 2001--2005; CADEL: mostly 2022--2023);
(3) the entire data, including the source texts, can be made publicly available; and
(4) it is linked to multilingual knowledge by adopting Wikidata.

\subsection{Major Studies on EL Design Choices} \label{sec:design}
EL may naively be defined as ``the task of identifying, for a linguistic expression, the KB entry that \textit{corresponds} to the concept it denotes.''
However, it is not trivial in which cases a correspondence relation between a linguistic expression and an entry should be recognized.
In particular, difficult cases arise when there exists an entry that is not directly identical to the concept referred to by the mention, but is nevertheless related to it, in relations such as whole--part or predecessor--successor relations.
Accordingly, criteria are needed not only for mentions but also for entries.

\paragraph*{Ling et al.'s Design Choices.}
\citet{ling-etal-2015-design} pointed out that no standard task definition for EL had been established, and identified the following five items (perspectives) as design choices that should be defined in annotation guidelines to ensure data consistency:
(L1) Target entity type, (L2) Overlapping mentions, (L3) Common noun mentions, (L4) Metonymic mentions, and (L5) Granularity of linked entities.\footnote{The identifiers L1--L5 (and R1--R4 introduced later) are assigned in this paper for convenience of reference.}
With respect to L1--L4, various policies are possible depending on the purpose of corpus construction and constraints such as cost. L5 involves a composite perspective and leaves room for further subdivision into more specific situations.

\paragraph*{Rosales-M\'{e}ndez et al.'s Categorization Scheme.}
\citet{rosales-mendez-etal-2019-fine} conducted a questionnaire survey of EL researchers based on example sentences asking what the output of an ideal EL system should be, and showed that agreement among respondents (36 in total) varied substantially across different dimensions.\footnote{For example, from the perspective of entity type, ``Living with Michael Jackson'' (a program title) showed high agreement, with 97\% of respondents selecting it as a linking target. By contrast, from the perspective of mention expression type, ``he and his four siblings'' was selected by 50\% of respondents, while ``he'' within the same expression was selected by 56\%, indicating low agreement.}
They concluded that there is no universal task definition suitable for all situations, and that the criteria to be adopted depend on the intended application.
They then proposed a categorization scheme consisting of the following four major categories: (R1) Base form of the mention, (R2) Part of Speech (POS) of the mention, (R3) Overlap between mentions, and (R4) Referential relation between the mention and its linked entity.
Furthermore, they suggested defining the EL task by deciding which subcategories to include as annotation targets.

\subsection{Design Choices Adopted in Japanese EL Datasets} \label{sec:design_ja}
\citet{jargalsaikhan-etal-2016-building} adopt the criterion: ``Choose the entity that is the most specific in possible candidates'' and illustrate it with several examples, including assigning $\langle$2002 FIFA World Cup$\rangle$ to the mention ``World Cup.''
However, which entry should be selected as the ``most specific'' appears to be left to the annotator's judgment.

\citet{murawaki-mori-2016-wikification} adopt a criterion that permits ``Topical Matching''---that is, recognizing a match on the basis of topical overlap---and state that this includes mention--entity pairs such as those involving hypernym--hyponym relations, an organization and its predecessor, and near-synonymous relations such as ``manufacturer'' and ``manufacturing industry.''

In the Shinra linking task definition,\footnote{\url{https://drive.google.com/file/d/1PpaYRBvxrrqUPNuAC48l1h6fzIiyaxAM/view}} the degree of match between a mention (attribute value) and an entry (Wikipedia article) is divided into ``exact match'' and ``partial match,'' and three types of cases are listed as falling under partial match: (1) Later Name, (2) Part-of, and (3) Derivation-of.
Case (1) includes examples such as assigning the entry $\langle$AGC$\rangle$, the company's later name, to the mention ``Asahi Glass.''
Case (2) includes examples such as assigning the entry $\langle$Tokyo Woman's Christian University$\rangle$, the superordinate organization, to the mention ``School of Arts and Sciences, Tokyo Woman's Christian University.''
Case (3) includes examples such as assigning the entry $\langle$A Midsummer Night's Dream$\rangle$, the name of the literary work, to the mention ``Midsummer Night's Dream,'' referring to a stage performance.

\paragraph*{Summary}
Design choices concerning entity types and mentions can be determined according to the intended purpose.
By contrast, non-trivial issues remain in entry selection.
Difficult cases of entry selection such as those discussed above can be understood as cases of conceptual overlap between the referent of a mention and available KB entries.
These cases arise when the same or similar names are used for concepts whose boundaries do not fully coincide.
They are further complicated by the uneven and incomplete nature of KB, which limits the entries actually available for selection.
Although this issue has been discussed in previous studies, including item L5 of \citet{ling-etal-2015-design}, existing studies still lack comprehensive coverage in terms of what kinds of conceptual overlap should be permitted as link targets.

\section{Design Policy for CADEL} \label{sec:policy}

\begin{table*}[t]
\centering
\small
\begin{tabular}{lc|lc|lc|lc|lc}
\toprule
\multicolumn{8}{c|}{Mention}  & \multicolumn{2}{c}{Entry} \\
\midrule
\multicolumn{2}{c|}{(R1)~Base form} & \multicolumn{2}{c|}{(R2)~POS} & \multicolumn{2}{c|}{(R3)~Overlap} & \multicolumn{2}{c|}{(R4)~Reference}  & \multicolumn{2}{c}{Link type} \\
\midrule
PROPN: Full name & $\checkmark$ &NP: Singular & $\checkmark$ & None &  $\checkmark$ 
& Direct &  $\checkmark$ & Exact &  $\checkmark$ \\
PROPN: Short name      & $\checkmark$ & NP: Plural & $\checkmark$ & Maximal &  $\checkmark$
& Anaphoric & $\triangle$ & Related &  $\checkmark$ \\
PROPN: Extended name & $\checkmark$ & Adjective     &              & Intermediate    &
& Metaphoric & & -- \\
PROPN: Alias      & $\checkmark$ & Verb       &              & Minimal &
& Metonymic & $\checkmark$ & -- \\
Numeric/Temporal     &              & Adverb     &              & -- &
& Related & & -- \\
Common form          & $\triangle$  & --       &              & -- &
& Descriptive & $\checkmark$ & -- \\
Pronoun form            & $\triangle$  & --       &              & -- &
& -- & & -- \\
\bottomrule
\end{tabular}
\caption{Overview of CADEL's design policy, which extends~\citet{rosales-mendez-etal-2019-fine}'s taxonomy (R1--R4). The subcategory ``Related'' under R4 follows their taxonomy and is distinct from our ``Related'' entry link type. PROPN and NP indicate proper noun and noun phrase, respectively.}
\label{tab:cadel_taxonomy}
\end{table*}

In this section, we present our design policies for entity types, mentions, and entries; the corresponding annotation criteria are provided in Appendix~\ref{sec:anno_criteria}.
In particular, we distinguish cases of conceptual overlap from ordinary ``exact links'' by treating them as ``related links'' whose types are further classified (\S\ref{sec:criteria_entry}).
As the KB, we adopt Wikidata,\footnote{\url{https://www.wikidata.org/}} a structured database whose sources are mainly multilingual Wikipedia articles.

\begin{table}[t]
\centering
\small{
\begin{tabular}{lll} \toprule
Tag & Type & ENE category \\ \midrule
\texttt{PER} & Person   & 1.1 \\
\texttt{LIV} & Living things & 1.3 \\
\texttt{ORG} & Organization & 1.4 \\
\texttt{LOC} & Location     & 1.5 \\
\texttt{FAC} & Facility    & 1.6 \\
\texttt{PRO} & Product & 1.7 (subset)\\
\texttt{EVE} & Event & 1.8 \\
\texttt{TIME} & Time  & 2.1 (proper names only) \\
\texttt{NOMINAL} & -- & -- \\
\bottomrule
\end{tabular}
}
\caption{Entity types for CADEL. \texttt{NOMINAL} is used for non-named mentions across entity types.} \label{tab:category}
\end{table}

\subsection{Entity Types}
The most fundamental design choice concerns the types of entities, corresponding to item L1 of \citet{ling-etal-2015-design}.
\citet{rosales-2018-should} argue that the question to be resolved in EL is not ``What is an entity?'' but rather ``What should be linked in EL?'', and that the answer depends on the target application.
Our goal in this study is to construct a basic EL dataset that can be used independently of any specific downstream application.
Accordingly, we define entities by balancing two considerations: broad coverage of the types typically targeted in domain-independent information extraction, and the ease of determining the appropriate entry to assign.
Specifically, we treat the following as entities:
\textit{concrete and abstract referents that have proper names and can be identified through linguistic expressions}.\footnote{It is not necessary for a corresponding entry to actually exist in the KB.}

More concretely, as shown in Table~\ref{tab:category}, we adopt the following types based on the second level of the ENE hierarchy Version 9.\footnote{\url{http://ene-project.info/ene9/}}
Among the categories corresponding to ``1.~Name'' in the ENE hierarchy, we target those that correspond to the entities defined above: person names, names of living things, organization names, location names, facility names, a subset of product names,
and event names.
In addition, among the categories corresponding to ``2.~Temporal Expression,'' we also include dates with proper names (e.g., ``Children's Day'') and named historical periods (e.g., ``the Kamakura period'').
We exclude names of natural objects (such as substance names and species names), disease names, color names, and some subclasses of product names (such as title names and unit names), since these are considered to fall outside the prototypical cases of entities defined above.

\subsection{Mention Identification}\label{sec:policy_mention}
The classification criteria R1--R4 of \citet{rosales-mendez-etal-2019-fine} are generally concerned with mention identification\footnote{R1 includes L3 (of Ling et al.); R3 corresponds to a reorganization of L2; and R4 partially overlaps with L4.}.
Our policy for these criteria is defined as follows and summarized in Table~\ref{tab:cadel_taxonomy}.

\paragraph*{R1: Base form of mentions.}
For proper names, we include informal names and abbreviations.
For common noun phrases and demonstrative pronouns, we include only cases where the referent can be identified from the context, such as when they corefer with a proper name in the document (for reasons of usefulness and annotation cost, we do not attempt exhaustive annotation; accordingly, Table~\ref{tab:cadel_taxonomy} marks this item with ``$\triangle$'').
Numerical and temporal expressions are excluded because they fall outside the entity types covered by this corpus.

\paragraph*{R2: Part of speech of mentions.}
We primarily target noun phrases and pronouns that refer to entities.
However, this restriction does not apply when a non-noun expression is used as the name of an entity.

\paragraph*{R3: Overlap among mentions.}
For nested linguistic expressions, we treat only the outermost expression as a mention, provided that it corresponds to one of the entity types covered by this study.
The main reason is that multilayer annotation over nested structures imposes a substantial annotation burden.

\paragraph*{R4: Referential relation.}
In principle, we determine whether a linguistic expression should be treated as a mention on the basis of its form and referential content.
Accordingly, for expressions involving direct, anaphoric, metonymic, or descriptive reference, we assign the entry corresponding to the referent as the link target.
By contrast, we do not cover linking between common nouns based on relations such as near-synonymy, hypernymy, and hyponymy (corresponding to the subcategory ``Related'' in \citet{rosales-mendez-etal-2019-fine}).

\subsection{Entry Assignment} \label{sec:policy_entry}
In this study, we adopt the following policy for assigning entries.
(a) An \textit{exact link}, or \textit{linking by exact match}, is assigned when the KB contains an entry corresponding to the concept referred to by the mention.
(b) A \textit{related link}, or \textit{linking by relatedness}, is assigned in cases of conceptual overlap when no such entry exists in the KB.
In such cases, the closest available entry among those that conceptually overlap is assigned; only when no such entry exists is no link target assigned.
We further distinguish types of related links by tags.

This is similar to the policy adopted in the Shinra linking task, but we introduce a more comprehensive classification of related links in \S\ref{sec:anno_link} and Appendix~\ref{sec:criteria_entry}.

\section{Corpus Construction Procedure} \label{sec:pre_anno}
We constructed CADEL in four steps: (1) document selection, (2) text preprocessing and article selection, (3) mention annotation, and (4) coreference and link annotation.
These steps were carried out by three annotators, all native speakers of Japanese, under the supervision of an annotation manager at a data annotation company.
The annotations in Steps 3 and 4 were performed using an annotation tool developed by the same company in Microsoft Access VBA.

\subsection{Document Selection}
As data sources, we manually selected HTML or PDF documents published on the websites of Japanese government ministries and agencies and their internal departments.\footnote{\url{https://www.gov-online.go.jp/topics/link/index.html}.}
We chose these sites for two reasons: (1) they permit flexible reuse under terms of use compatible with the copyright license conditions of CC BY 4.0;\footnote{Terms of use for content on ``Government Public Relations Online'': \url{https://www.gov-online.go.jp/etc/tos.html}.} and
(2) they contain many documents describing or reporting on regions, events, policies, and history in Japan, making them a rich source of mention instances referring to entities specific to Japan.

\subsection{Text Preprocessing and Article Selection}
From each selected source file, we manually extracted all or part of the text and converted it into an XML format with the hierarchical structure ``document (doc) $\supset$ article (subdoc) $\supset$ section,'' defined on the basis of semantic 
content rather than the original HTML structure.
Here, a document corresponds to a single source file, an article to an independently readable content unit within that file, and a section to a smaller unit within an article.
The extracted text consisted of document titles, article titles, section headings, body text within sections (split sentence by sentence), and optionally other text within sections (e.g., headers and figure captions), each distinguished by dedicated XML tags we defined.
An example of a formatted document is shown in Figure~\ref{fig:sample_xml} in Appendix~\ref{sec:app_xml}.

\subsection{Mention Annotation}
As a preprocessing step, we applied GiNZA\footnote{\url{https://github.com/megagonlabs/ginza}}~\cite{ginza-2019} to the text inside the XML tags of each document to assign ENE tags, and then converted them into the entity type tags used in our corpus.
Annotators then identified mentions in the documents (by correcting and supplementing the automatically assigned results) and assigned entity type tags.
The detailed annotation criteria is described in Appendix~\ref{sec:men_anno}.

\subsection{Coreference and Link Annotation} \label{sec:anno_link}
Annotators grouped mentions within each article into coreference clusters when they could be interpreted as referring to the same entity, and assigned Wikidata entry IDs corresponding to their referents at the cluster level, by correcting the automatically assigned results based on exact matches between mentions and entry names.\footnote{Clusters were not merged across articles. However, it is possible to aggregate clusters assigned the same entry ID.}
For each mention, when there was no exact-link entry (i.e., no Wikidata entry directly corresponding to the referent), annotators assigned a related-link entry ID together with a tag indicating the type of related link.
When no appropriate entry could be found, the value ``\texttt{NIL}'' was assigned instead of an entry ID.
Although multiple entries may qualify as related links for a mention, we annotated only one entry due to annotation costs.

Specifically, based on annotation work on actual texts, we introduce the following types of related links:
\begin{enumerate}
\setlength{\itemsep}{0pt}
\setlength{\parskip}{0pt}
\setlength{\parsep}{0pt}
\item Inclusive and non-inclusive overlap: cases in which two entities stand in a whole--part relation or partially overlap,
\item Periodic series and instances: cases involving recurring events or related organizations, where both individual instances and the series as a whole can be referred to as entities,
\item Diachronic correspondence: cases in which the name or identity of an entity changes over time, resulting in a predecessor--successor relation,
\item Other conceptual overlap, and
\item Indeterminate concept definition: cases in which the concept referred to by the writer is itself vague.
\end{enumerate}
These types are defined and illustrated in Appendix~\ref{sec:criteria_entry}.

\section{Corpus Statistics and Characteristics}

As a result of the annotation process, we constructed a corpus consisting of 85 documents, 160 articles, and 3,850 sentences in total, as shown in Table~\ref{tab:basic_stat}.
Public relations magazines and white/blue papers account for 79\% of all articles, likely because their texts tend to be sufficiently long and contain a wide variety of entities.
Three source websites---those of the Ministry of Land, Infrastructure, Transport and Tourism, the Ministry of Agriculture, Forestry and Fisheries, and the Ministry of Defense (including their internal departments)---account for 66\% of all articles.
Documents published in 2022--2023 account for 85\% of all articles.\footnote{The Wikidata entries assigned during annotation were limited to those that existed at the time of annotation (up to February 2024).}
Among the 5,760 named-mention instances\footnote{This is the sum of 5,058 exact links and 702 related links in the \texttt{NAME} row of Table~\ref{tab:men_stat_each_type}.} (1,799 distinct mentions) that were assigned Wikidata entries, 87\% of the assigned entries (80\% in terms of distinct entries) have a Country property, and among those, 87\% (85\% in terms of distinct entries) have \texttt{Q17} (Japan) as the value of that property.
These figures also confirm that the corpus contains many proper names specific to Japan.
More detailed statistics are provided in Appendix~\ref{sec:app_stat}.

\paragraph*{Annotation information.}
For the annotations in Step 3 (mentions) and Step 4 (coreference and links), 150 articles within 80 documents were annotated by a single annotator each.
The remaining 10 articles within 5 documents were independently annotated by two annotators each, and inter-annotator agreement was measured for them, as described later in \S\ref{sec:iaa}.
For these 10 articles, annotation inconsistencies were resolved through discussion involving the first author and the annotation manager, and the revised annotations were used as the final data.
As a result of annotation, the corpus contains 8,082 mention instances in total (3,669 distinct mentions), of which 6,462\footnote{This is the sum of 5,569 exact links and 893 related links in the ``All'' row of Table~\ref{tab:men_stat_each_type}.} (2,594 distinct mentions) were assigned a Wikidata entry ID as either an exact or related link, while 1,620 mention instances (1,103 distinct mentions) were assigned \texttt{NIL}.

\begin{table}[t]
\centering
\small
\begin{tabular}{ccccc}
\toprule
\#Doc & \#Article & \#Sent & \#Mention & \#Cluster\\
\midrule
85 & 160 & 3,852 & 8,082 & 4,049 \\
\bottomrule
\end{tabular}
\caption{Basic statistics of CADEL.}
\label{tab:basic_stat}
\end{table}

\begin{table}[t]
\centering
\small
\begin{tabular}{lrrrrr}
\toprule
Type & Mention & \multicolumn{2}{c}{Exact link (\%)} & \multicolumn{2}{c}{Related link (\%)}\\
\midrule
\texttt{NAME}      & 6,939 & 5,058 & (72.9) & 702 & (10.1) \\
\quad \texttt{PER} &   307 &   217 & (70.7) &   1 & (0.0) \\ 
\quad \texttt{LIV} &     8 &     0 & (0.0)  &   0 & (0.0) \\ 
\quad \texttt{ORG} & 1,319 &   957 & (72.6) & 156 & (11.8) \\ 
\quad \texttt{LOC} & 2,596 & 2,228 & (85.8) & 297 & (11.4) \\ 
\quad \texttt{FAC} & 1,059 &   670 & (63.3) & 173 & (16.3) \\
\quad \texttt{PRO} &   673 &   278 & (41.3) &  53 &  (7.8) \\
\quad \texttt{EVE} &   471 &   179 & (38.0) &  60 & (12.7) \\
\quad \texttt{TIME}&   506 &   492 & (97.2) &   0 & (0.0) \\
\texttt{NOMINAL}\!\!\!\! & 1,143 & 548 & (47.9) & 154 & (13.5) \\
\midrule
All & 8,082 & 5,569 & (68.9) & 893 & (11.0) \\
\bottomrule
\end{tabular}
\caption{The numbers of mentions and link numbers/ratios of Wikidata entries for each type.}
\label{tab:men_stat_each_type}
\label{tab:cate_dist}
\end{table}

\paragraph*{Number of mentions and link rates}
Table~\ref{tab:men_stat_each_type} shows, for each entity type, the total number of mention instances and the number and proportion (link rate) of mentions assigned Wikidata entries as either exact or related links.
Location names \texttt{LOC}, facility names \texttt{FAC}, and organization names \texttt{ORG} are central types in this corpus, in that they occur frequently and also have relatively high link rates.
Across all types, about 70\% of mentions receive exact links, and about 80\% receive either exact or related links.
Therefore, a task setting that includes related links would increase the proportion of linkable mentions and is likely to help develop useful EL systems.

\section{Analysis}
\subsection{Inter-Annotator Agreement} \label{sec:iaa}
Based on the annotations produced independently by two annotators for the same set of 10 documents, we measured inter-annotator agreement for mention, coreference, and link annotation.

For mention annotation, we measured F1 score.
Agreement was high for proper names, reaching 0.825--0.846 for exact match of both span and entity type.
By contrast, agreement was lower for non-named mentions (\texttt{NOMINAL}), at 0.522--0.547.

\begin{table*}[t]
\small
\centering
\begin{tabular}{lc|ccc|ccc}
\toprule
\multirow{2}{*}{Ranking method} & \multirow{2}{*}{\texttt{AltLabel}} & \multicolumn{3}{c|}{Exact link} & \multicolumn{3}{c}{Related link}\\
&& \multicolumn{1}{c}{R@1} & \multicolumn{1}{c}{R@10} & \multicolumn{1}{c|}{R@100} & \multicolumn{1}{c}{R@1} & \multicolumn{1}{c}{R@10} & \multicolumn{1}{c}{R@100} \\
\midrule
\multirow{2}{*}{Uniform}
& & 0.555 & 0.756 & 0.762 & 0.074 & 0.099 & 0.100 \\
& $\checkmark$ & 0.565 & 0.788 & \textbf{0.795} & 0.086 & 0.112 & \textbf{0.113} \\
\midrule
\multirow{2}{*}{SmallerID}
& & 0.693 & 0.761 & 0.762 & 0.080 & 0.100 & 0.100 \\
& $\checkmark$ & 0.716 & 0.794 & \textbf{0.795} & 0.091 & \textbf{0.113} & \textbf{0.113} \\
\midrule
\multirow{2}{*}{MoreWikiBLs}
& &
0.733 & 0.761 & 0.762 & 0.098 & 0.100 & 0.100 \\
& $\checkmark$ &
\textbf{0.755} & \textbf{0.795} & \textbf{0.795} & \textbf{0.110} & \textbf{0.113} & \textbf{0.113} \\
\bottomrule
\end{tabular}
\caption{Disambiguation performance of string matching for named mentions.} 
\label{tab:spqrql_result}
\end{table*}

For coreference annotation, we measured agreement using LEA~\cite{moosavi-strube-2016-coreference} and the CoNLL score~\cite{pradhan-etal-2012-conll}, both metrics for coreference resolution, and obtained high agreement, with F1 scores above 0.9 on both metrics.

For link annotation at the mention level, we measured agreement using both F1 score and Cohen's $\kappa$.
Agreement was high (F1 = 0.830--0.924, $\kappa$ = 0.817--0.919) when evaluation was restricted to exact links, whereas it was somewhat lower when both exact and related links were taken into account.

Taken together, the results indicate a high level of annotation consistency in the corpus.

\subsection{Preliminary Evaluation by String Matching} \label{sec:pre_eval}
Using this corpus, we conducted a preliminary entity disambiguation experiment based on string matching.
Specifically, for each named mention assigned either an exact link or a related link, we retrieved a set of candidate entries by issuing a query containing the normalized string\footnote{Starting from the original string normalized by NFKC, we further applied normalization such as removing parentheses ``()'' and the enclosed substring.} of the gold mention through the Wikidata SPARQL Query Service\footnote{Web interface: \url{https://query.wikidata.org/}} (Figure~\ref{fig:sparql_query} in Appendix~\ref{sec:exp_detail}), and then checked whether the gold entry was included in the retrieved set.\footnote{Entries corresponding to disambiguation pages were excluded from the retrieved candidate set.}
As the evaluation metric, we used the expected value of Recall@$k$ ($k \in \{1,10,100\}$), i.e., the proportion of cases in which the correct entry is included among the top-$k$ outputs (see Appendix~\ref{sec:exp_detail} for the calculation details).

To rank the retrieved candidate set, we used three methods:
(1) \textsc{Uniform}, which assigns the same score to all entries;
(2) \textsc{SmallerID}, which ranks entries in ascending order of the numeric part of the Wikidata entry ID, under the assumption that smaller IDs tend to correspond to more popular entries; and
(3) \textsc{MoreWikiBLs}, which ranks entries in descending order of the number of backlinks to the corresponding Japanese Wikipedia article.\footnote{Backlink counts were obtained using the MediaWiki API (\url{https://www.mediawiki.org/wiki/API:Main_page/}). Note that the number of backlinks returned by the API is capped at 500.}
For each of these three methods, we also compared two candidate-retrieval settings:
(a) retrieving only entries whose label exactly matches the normalized mention string, and
(b) retrieving entries whose label or alternative label (\texttt{AltLabel}) exactly matches the normalized mention string.

Table~\ref{tab:spqrql_result} shows the results.
For exact-link cases, the naive method \textsc{Uniform} achieved Recall@1 of 0.555, indicating that more than half of the cases can be resolved correctly by exact string matching alone.
The two popularity-based heuristics, \textsc{SmallerID} and \textsc{MoreWikiBLs}, achieved Recall@1 values of 0.693--0.755, showing that such popularity-based heuristics are effective for disambiguation.
In addition, across all three methods, using alternative labels improved Recall at each $k$ by approximately 0.01--0.03 points.
Because the maximum size of the retrieved candidate set was around 80, the value of Recall@100 was identical across ranking methods.

For related-link cases, the two popularity-based heuristics yielded only slight improvements, and the Recall@$k$ values indicate that the correct entry was retrieved in only around 10\% of the cases.
This is unsurprising, since related-link entries based on relations such as whole--part or predecessor--successor are often expected to have names that differ from those of the mention's direct referent.
Given the large performance gap between exact-link and related-link cases, evaluating them separately  is likely to be useful for understanding system performance in detail.

\subsection{Discussion} \label{sec:discussion}
\paragraph*{Data Split for Evaluation}

The results on exact-link cases (with a maximum Recall@1 of 0.755) indicate that approximately 75\% of all named mentions can be disambiguated by simple methods, suggesting that the corpus as a whole is skewed toward easy cases.
This is likely due to the nature of the source documents, which are published by government institutions and therefore tend to use official and widely accepted surface forms.
In contrast, performance on related-link cases was much lower (around 10\%), indicating that these cases are substantially more difficult.
Taken together, these results suggest that while many exact-link cases can be resolved by simple methods, a substantial number of exact-link cases (about 1,240 mention instances), as well as most related-link cases remain challenging.
Based on this result, we created an official data split (Table~\ref{tab:data_split}) in which articles were preferentially assigned to the test set if they contained many mentions whose gold entries could not be identified by the \textsc{SmallerID} + \texttt{AltLabel} method, along with other articles from the same source documents.
This split is intended to make the corpus a more useful evaluation benchmark for distinguishing differences in EL system performance.

\begin{table}[t]
\centering
\small
\begin{tabular}{lrrrr}
\toprule
 & \#Doc & \#Art & \#Sent & \#Mention \\ \midrule
Train & 39 & 74 & 1,607 & 2,903 \\
Dev   & 15 & 31 &   636 & 1,354 \\
Test  & 31 & 55 & 1,609 & 3,825 \\ \midrule
Total & 85 & 160 & 3,852 & 8,082 \\
\bottomrule
\end{tabular}
\caption{Official data split of CADEL.}
\label{tab:data_split}
\end{table}

\paragraph*{Additional Resources}
We also release another EL dataset, EnJaEL,\footnote{\url{https://github.com/shigashiyama/en-ja-el}}, which consists of Japanese translations of three English EL datasets: VoxEL~\cite{rosales-2018-voxel}, MEANTIME~\cite{minard-etal-2016-meantime}, and Linked-DocRed~\cite{tan-etal-2022-revisiting}.
This dataset was created from the original datasets using machine translation with full post-editing and annotation projection, and serves as complementary resources for evaluation on topics and concepts originating outside the Japanese context.

\paragraph*{Subsequent Use}
As a notable study published after the public release of CADEL in January 2025, \citet{sawada-2026} compared the performance of state-of-the-art multilingual entity disambiguation methods using several Japanese EL corpora, including CADEL.
On the CADEL test set, they reported disambiguation accuracy of up to 62.4 in Recall@1 and 83.3 in Recall@10 over both exact and related links, which were the second lowest scores among the five corpora used in their study.
Their experiments provide a useful comparison of representative disambiguation methods on this corpus.
Nevertheless, further analysis using CADEL would enable more detailed system evaluation, including end-to-end evaluation that incorporates mention recognition and separate evaluation of disambiguation accuracy for exact-link and related-link cases.

\section{Conclusion}
In this paper, we described the design policy and construction procedure of CADEL, a corpus of administrative web documents for Japanese entity linking.
We also analyzed the characteristics of the corpus through inter-annotator agreement evaluation and a preliminary disambiguation experiment based on string matching, and discussed its potential uses.
We hope that this study will contribute to further advances in Japanese EL research, both in language resources and in system development.

\bibliography{custom}

\clearpage
\appendix

\section{Annotation Criteria} \label{sec:anno_criteria}
\subsection{Criteria for Mention Identification} \label{sec:men_anno}
Proper names belonging to any of the types from \texttt{PER} to \texttt{TIME} in Table~\ref{tab:category} were annotated as mentions and assigned the corresponding type tags.
Non-named expressions, such as common noun phrases and demonstrative expressions, were annotated as mentions and assigned the tag \texttt{NOMINAL} when their referents could be identified from context.

Mention identification, including span boundaries, followed the criteria below:
\begin{enumerate}[topsep=2pt,parsep=0pt]
\item In nested structures, only the outermost expression is annotated as a mention (\S\ref{sec:policy_mention}).
\item To improve consistency in annotation units, expressions that can be interpreted as a sequence of multiple mentions are split into multiple mentions.
However, when the entire expression can be regarded as an official name or as referring to a single entity, it is annotated as a single mention.\footnote{Addresses were split as in (c). Place names smaller than municipalities were not split, because their boundaries are difficult to determine.}

\begin{enumerate}
    \item \uline{Ito En}${}^{\scriptsize\texttt{ORG}}$~\uline{Healthy Mineral Mugicha}${}^{\scriptsize\texttt{PRO}}$
    \item \uline{Kyoto City Zoo}${}^{\scriptsize\texttt{FAC}}$
    \item \uline{Kyoto Prefecture}${}^{\scriptsize\texttt{LOC}}$~\uline{Seika Town}${}^{\scriptsize\texttt{LOC}}$~\\\uline{Hikaridai 3-5}${}^{\scriptsize\texttt{LOC}}$
    \end{enumerate}
\item To avoid redundant annotation, we do not annotate attributive information appearing in appositional constructions or copular sentences.
However, when the entire expression, including an appositional modifier, can be regarded as an official name, it is annotated as a single mention.
    \begin{enumerate}
    \setcounter{enumii}{3}
    \item the northernmost station in Japan ``\uline{Wakkanai Station}${}^{\scriptsize\texttt{FAC}}$''
    \item \uline{Michi-no-Eki Hari T.R.S.\ (Hariterasu)}${}^{\scriptsize\texttt{FAC}}$
    \end{enumerate}
\end{enumerate}

\begin{figure*}[t]
\small
\begin{quote}
<?xml version="1.0" encoding="UTF-8"?>\\
<doc date="20220113" src="https://www.maff.go.jp/j/press/kanbo/anpo/230113.html">\\
\hspace*{2em}<title>\ja{「食から日本を考える。NIPPON FOOD SHIFT FES.山梨」を開催！}</title>\\
\hspace*{2em}<subdoc>\\
\hspace*{4em}<section>\\
\hspace*{6em}<heading>\ja{～～八ヶ岳から、ニッポンフードシフト。～～}</heading>\\
\hspace*{6em}<line>\ja{農林水産省では、食と農のつながりの深化に着目した国民運動「食から日本を考える。ニッポンフードシフト」を実施しています。}</line>\\
\hspace*{6em}...\\
\hspace*{4em}</section>\\
\hspace*{2em}</subdoc>\\
</doc>\\
\end{quote}
\caption{Example of an XML-structured document.}  \label{fig:sample_xml}
\end{figure*}

\subsection{Types of Related Links} \label{sec:criteria_entry}
For related links, we define eight tags in the following five categories to distinguish the relation between the mention and the assigned entry.
No such tag is assigned for exact links.
\begin{enumerate}[topsep=2pt,parsep=0pt]
\item Inclusive and non-inclusive overlap: cases in which two entities stand in a whole--part relation or partially overlap.
\begin{itemize}
    \item When a mention (e.g., ``\uline{Faculty of} \uline{Education}, Nara University of Education'') is assigned an entry that includes it as a part (e.g., $\langle$Nara University of Education{${}^{\scriptsize\texttt{Q1629024}}$}$\rangle$), we assign the tag \texttt{PART\_OF}.
    \item When a mention (e.g., ``the three major pine groves of Japan'') is assigned an entry corresponding to one of its parts (e.g., $\langle$Miho no Matsubara{${}^{\scriptsize\texttt{Q869275}}$}$\rangle$), we assign the tag \texttt{CONTAINS}.
    \item When a mention (e.g., ``Koyasan and its \uline{foothill area}'') is assigned an entry that includes a region partially overlapping with it, but whose inclusion relation is unclear (e.g., $\langle$Mount K\=oya{${}^{\scriptsize\texttt{Q535065}}$}$\rangle$), we assign the tag \texttt{SHARE}.
\end{itemize}

\item Periodic series and instances: cases involving recurring events or related organizations, where both individual instances and the series as a whole can be referred to as entities.
\begin{itemize}
    \item When a mention (e.g., ``Tsukuba Science Festival 2022'') is assigned an entry corresponding to its series (e.g., $\langle$Tsukuba Science Festival{${}^{\scriptsize\texttt{Q11272282}}$}$\rangle$), we assign the tag \texttt{PERIODIC\_INSTANCE\_OF}.
    \item Conversely, when a mention denoting a series is assigned an entry corresponding to one of its instances, we assign the tag \texttt{PERIODIC\_SERIES\_OF}.
\end{itemize}

\item Diachronic correspondence: cases in which the name or identity of an entity changes over time, resulting in a predecessor--successor relation.
\begin{itemize}
    \item When a mention (e.g., the former ``Tokyo Electric Power Co., Ltd.'' or the former ``Kamaishi City,'' whose municipal area differed from the current one) is assigned an entry corresponding to its successor (e.g., $\langle$Tokyo Electric Power Company Holdings{${}^{\scriptsize\texttt{Q333894}}$}$\rangle$ or the current $\langle$Kamaishi{${}^{\scriptsize\texttt{Q329790}}$}$\rangle$), we assign the tag \texttt{DIACHRONIC\_CORRESPONDENCE}.
    The same tag is also used when a mention is assigned an entry corresponding to its predecessor.
\end{itemize}

\item Other conceptual overlap
\begin{itemize}
    \item When a mention (e.g., ``Kiki's Delivery Service,'' referring to the 2021 stage musical) is assigned an entry that has conceptual overlap other than those in (1)--(3) (e.g., the children's novel by Eiko Kadono, $\langle$Kiki's Delivery Service{${}^{\scriptsize\texttt{Q1768944}}$}$\rangle$; the 1989 animated film by Hayao Miyazaki, $\langle$Kiki's Delivery Service{${}^{\scriptsize\texttt{Q196602}}$}$\rangle$; or the 2014 live-action film by Takashi Shimizu, $\langle$Kiki's Delivery Service{${}^{\scriptsize\texttt{Q196602}}$}$\rangle$), we assign the tag \texttt{OTHER}.
\end{itemize}

\item Indeterminate concept definition: cases in which the concept referred to in the text is itself vague, or it is unclear which of several official definitions applies.\footnote{This is distinct from ambiguity in which the same expression may refer to multiple different referents.}
\begin{itemize}
    \item When the precise referent of a mention (e.g., ``the feeling that I am in \uline{Kyoto}'') is unclear, we assign one plausible candidate (e.g., $\langle$Kyoto Prefecture{${}^{\scriptsize\texttt{Q120730}}$}$\rangle$, $\langle$Kyoto City{${}^{\scriptsize\texttt{Q34600}}$}$\rangle$, or $\langle$Kyoto{${}^{\scriptsize\texttt{Q740246}}$}$\rangle$ as a historical city or place name) mark it with the tag \texttt{VAGUE}.
\end{itemize}
\end{enumerate}

\section{Example Document in CADEL} \label{sec:app_xml}
Figure~\ref{fig:sample_xml} shows an example of a formatted XML document included in the corpus.
The text is quoted from the URL shown in the figure. Although the original page at that URL is no longer accessible, it is available on the archived web page.\footnote{\url{https://warp.ndl.go.jp/en/web/20230406001247/https://www.maff.go.jp/j/press/kanbo/anpo/230113.html}}

\section{Detailed Corpus Statistics} \label{sec:app_stat}
\paragraph*{Distribution of data sources}
Tables~\ref{tab:article_type}, \ref{tab:url_domain}, and \ref{tab:text_year} show the distributions of article media types, source website domains, and publication years in CADEL of the source documents, respectively.

\begin{table}[h!]
\centering
\small
\begin{tabular}{lrr}
\toprule
Text type & \#Docs & \#Articles \\
\midrule
Public relations magazines & 37 & 79 \\
White and blue papers & 21 & 48 \\
Press conference/meeting records & 7 & 10 \\
Press releases & 6 & 7 \\
Other web pages & 8 & 9 \\
Other PDF documents & 6 & 7 \\
\bottomrule
\end{tabular}
\caption{The numbers of documents/articles for each text media type.}
\label{tab:article_type}
\end{table}

\begin{table}[h!]
\centering
\small
\begin{tabular}{llrr}
\toprule
Org. & Domain &\#Docs & \#Articles \\
\midrule
CAO (Bousai) & \texttt{bousai.go.jp} & 4 & 4 \\
CAO (OGB) & \texttt{ogb.go.jp} & 1 & 2 \\
CAO (NPA) & \texttt{npa.go.jp} & 2 & 5 \\
MIC (FDMA) & \texttt{fdma.go.jp} & 1 & 2 \\
MOJ & \texttt{moj.go.jp} & 1 & 3 \\
MOFA & \texttt{mofa.go.jp} & 5 & 5 \\
MOF & \texttt{mof.go.jp} & 2 & 2 \\
MEXT & \texttt{mext.go.jp} & 2 & 7 \\
MEXT (ACA) & \texttt{bunka.go.jp} & 4 & 4 \\
MHLW & \texttt{mhlw.go.jp} & 1 & 1 \\
MAFF & \texttt{maff.go.jp} & 15 & 35 \\
METI & \texttt{meti.go.jp} & 8 & 11 \\
MLIT & \texttt{mlit.go.jp}& 24 & 41 \\
MLIT (GSI) & \texttt{gsi.go.jp} & 2 & 6 \\
ENV & \texttt{env.go.jp} & 2 & 3 \\
MOD & \texttt{mod.go.jp} & 11 & 29 \\
\bottomrule
\end{tabular}
\caption{The numbers of documents/articles for each source site domain name.} \label{tab:url_domain}
\end{table}

\begin{table}[h!]
\centering
\small
\begin{tabular}{lcc}
\toprule
Publication year & \#Docs & \#Articles \\
\midrule
Unknown & 6 & 7 \\
1998 & 1 & 1 \\
2017 & 1 & 1 \\
2018 & 1 & 1 \\
2019 & 1 & 5 \\
2020 & 3 & 4 \\
2021 & 4 & 5 \\
2022 & 22 & 39\\
2023 & 46 & 97\\
\bottomrule
\end{tabular}
\caption{The numbers of documents/articles for each publication year.}
\label{tab:text_year}
\end{table}

\paragraph*{Distribution of related links}
Table~\ref{tab:stat_linktype} shows the breakdown of all 893 mentions assigned related-link entries.
\texttt{PART\_OF} (\texttt{PART}) is the most frequent type, followed by \texttt{DIACHRONIC\_CORRESPONDENCE} (\texttt{DIAC}) and \texttt{OTHER} (\texttt{OTH}); together, these three account for 84.1\% of the total.
A smaller number of cases were assigned \texttt{CONTAINS} (\texttt{CONT}), \texttt{SHARE}, \texttt{PERIODIC\_INSTANCE\_OF} (\texttt{PRD\_I}), and \texttt{VAGUE}, while no cases were assigned \texttt{PERIODIC\_SERIES\_OF}.

\begin{table}[t]
\centering
\small
\begin{tabular}{ccc|c|c|c|c}
\toprule
\texttt{PART} & \texttt{CONT} & \texttt{SHARE} & \texttt{PRD\_I} & \texttt{DIAC} & \texttt{OTH} & \texttt{VAGUE}\\
\midrule
497 & 36 & 43 & 20 & 129 & 125 & 43\\
\bottomrule
\end{tabular}
\caption{Distribution of related link type tags assigned to mentions.}
\label{tab:stat_linktype}
\end{table}

\paragraph*{Characteristics of coreference clusters}
Table~\ref{tab:cls_stat} shows the number of coreference clusters by size.
Among all 4,049 clusters, 57.4\% are singletons (size 1), while the remaining 42.6\% have size 2 or larger.
Looking at the average number of distinct mentions within a cluster (DM: Distinct Mention) in Table~\ref{tab:cls_stat},\footnote{For example, suppose there are three clusters of size 3: $C_1=\{a^\star,b\}$, $C_2=\{c^\star,c^\star\}$, and $C_3=\{d,d\}$, where proper-name mentions are marked with ${}^\star$. Then $\textrm{DM}=(2+1+1)/3$ and $\textrm{DNM}=(1+1+0)/3$.} we find that the value increases with cluster size.
This indicates that the more often the same concept is mentioned within an article, the more likely it is to be referred to by different expressions.
By contrast, the number of distinct named mentions within a cluster (DNM: Distinct Named Mention) is shifted toward values close to 1, indicating that it is relatively rare for the same concept to be referred to by multiple different proper names.

\begin{table}[t]
\small
\centering
\begin{tabular}{l|rrrrrr}
\toprule
Size  &      1 &    2 &    3 &    4 &    5 & $\geq$6\\
\midrule
Clster  & 2,531 &  682 &  336 &  197 &   89 &  214 \\
DM   &  1.00 & 1.32 & 1.57 & 1.78 & 2.01 & 2.73 \\
DNM  &  0.95 & 1.13 & 1.24 & 1.38 & 1.42 & 1.72 \\
\bottomrule
\end{tabular}
\caption{The number of coreference clusters and the average number of mentions (DM: distinct mentions; DNM: distinct named mentions) inside clusters for each size.
The DM and DNM values for all clusters are 1.25 and 1.07, respectively.} \label{tab:cls_stat}
\end{table}

\section{Inter-Annotator Agreement} \label{sec:iaa_detail}
For the inter-annotator agreement study, among the 10 articles within 5 documents, 6 articles were annotated by annotators A and B, while the remaining 4 articles were annotated by annotators A and C; mention, coreference, and link annotations were conducted for these articles.
For each type of annotation, we report agreement scores for each annotator pair.
Because the coreference and link annotations were performed by the same annotator who had carried out the preceding mention annotation, the sets of mentions appearing in the same article are not identical between the two annotators in the latter stages.
Therefore, inter-annotator agreement for coreference and linking was measured only on mentions whose spans matched between the two annotators' results.\footnote{There were 315 such mentions for annotator pair A\&B and 147 for A\&C.}

\subsection{Mention Annotation}
Table~\ref{tab:iaa_mention} shows inter-annotator agreement for mention annotation, measured by F1 score.\footnote{The numbers of annotated mentions were as follows: for annotator pair A\&B, annotator A annotated 364 mentions (including 322 proper names) and annotator B annotated 344 mentions (including 291 proper names); for annotator pair A\&C, annotator A annotated 178 mentions (including 149 proper names) and annotator C annotated 161 mentions (including 144 proper names).}
Overall, there is no substantial difference in agreement between annotator pairs A\&B and A\&C.
A common tendency across both pairs is that proper names (``named'' in Table~\ref{tab:iaa_mention}) show high agreement (0.884--0.915 for exact span-only match and 0.825--0.846 for exact match for both span and entity type).
By contrast, non-named mentions (\texttt{NOMINAL}) show lower agreement (0.522--0.547), suggesting substantial individual variation in applying the criterion of whether the referent is identifiable.

\begin{table}[t]
\small
\centering
\begin{tabular}{l|cc|cc}
\toprule
\multirow{2}{*}{Mention}
& \multicolumn{2}{c|}{A vs. B} & \multicolumn{2}{c}{A vs. C} \\
& \multicolumn{1}{c}{Span} & \multicolumn{1}{c|}{+Type} & \multicolumn{1}{c}{Span} & \multicolumn{1}{c}{+Type} \\
\midrule
All   & 0.890 & 0.788 & 0.867 & 0.802 \\
Named  & 0.884 & 0.825 & 0.915 & 0.846  \\
Non-named & 0.547 & 0.547 & 0.522 & 0.522 \\
\bottomrule
\end{tabular}
\caption{Inter-annotator agreement (F1 score) for mention annotation. ``Span'' indicates exact span-only match, and ``+Type'' indicates exact match for both span and entity type.} 
\label{tab:iaa_mention}
\end{table}

\subsection{Coreference Annotation}
Table~\ref{tab:iaa_coref_cls} shows inter-annotator agreement for coreference annotation.
We used the F1 scores of LEA~\cite{moosavi-strube-2016-coreference} and the CoNLL score~\cite{pradhan-etal-2012-conll}, defined as the macro-average of MUC~\cite{vilain-etal-1995-model}, B${}^3$~\cite{bagga-etal-1998-algorithms}, and CEAF${}_e$~\cite{luo-2005-coreference}, as measures of inter-annotator agreement.

Overall, annotator pair A\&C shows slightly higher agreement.
Under both the setting that considers all mentions in a cluster and the setting that considers only named mentions within a cluster (excluding non-named mentions), both annotator pairs A\&B and A\&C achieve high agreement above 0.9 on all metrics.
Under the setting that considers only non-named mentions within a cluster, agreement ranges from 0.603 to 0.793 across metrics, indicating that these cases are more prone to variation than named mentions.

In addition, for the input mention set (i.e., all mentions whose spans matched between the two annotators), we created outputs using the Leave-as-is (LAI) method, which treats each mention as an independent cluster, and measured agreement between LAI and each annotator.
Of the two agreement scores between annotator X and LAI and between annotator Y and LAI, the higher one is reported in Table~\ref{tab:iaa_coref_cls} as ``(X or Y) vs.\ LAI.''
Agreement with LAI is low across all metrics and settings (0.286--0.510), suggesting that the annotators' outputs were obtained by appropriately performing coreference clustering based on the content of the text.

\begin{table}[t]
\small
\centering
\begin{tabular}{l|cc|cc}
\toprule
\multirow{2}{*}{Mention}
& \multicolumn{2}{c|}{A vs. B} & \multicolumn{2}{c}{(A or B) vs. LAI} \\
& \multicolumn{1}{c}{LEA} & \multicolumn{1}{c|}{CoNLL} & \multicolumn{1}{c}{LEA} & \multicolumn{1}{c}{CoNLL} \\
\midrule
All     & 0.906 & 0.939 & 0.470 & 0.475 \\
Named   & 0.909 & 0.926 & 0.512 & 0.496 \\
Non-named & 0.603 & 0.635 & 0.286 & 0.395 \\
\midrule
\midrule
\multirow{2}{*}{Mention}
& \multicolumn{2}{c|}{A vs. C} & \multicolumn{2}{c}{(A or C) vs. LAI}\\
& \multicolumn{1}{c}{LEA} & \multicolumn{1}{c|}{CoNLL} & \multicolumn{1}{c}{LEA} & \multicolumn{1}{c}{CoNLL} \\
\midrule
All     & 0.924 & 0.950 & 0.340 & 0.442 \\
Named   & 0.943 & 0.962 & 0.379 & 0.461 \\
Non-named & 0.720 & 0.793 & 0.462 & 0.510 \\
\bottomrule
\end{tabular}
\caption{Inter-annotator agreement for coreference annotation.}
\label{tab:iaa_coref_cls}
\end{table}

\subsection{Link Annotation}
Table~\ref{tab:iaa_link} shows inter-annotator agreement for link annotation at the mention level, measured by F1 score and Cohen's $\kappa$.
Under the ``exact'' setting, cases in which a Wikidata entry ID was assigned as an exact link were counted as InKB, whereas all other cases, including those assigned a related link type tag or \texttt{NIL}, were counted as Out-of-KB (OOKB), and agreement was calculated accordingly\footnote{For this calculation, cases with a related link type tag were treated as equivalent to \texttt{NIL}.}.
Under the ``exact \& related'' setting, cases were counted as InKB or OOKB based on whether the annotators agreed on the combination of entry ID and related link type tag (or \texttt{NIL}).

Annotator pair A\&B shows higher agreement, and both annotator pairs achieve generally high overall agreement (F1 = 0.769--0.924, $\kappa$ = 0.759--0.919).
In both annotator pairs, agreement is higher under the ``exact'' setting than under the ``exact \& related'' setting.
This is intuitive, given that candidates for related links are less likely to be uniquely determined.
Within the same setting, whether agreement was higher for InKB or for OOKB cases varied across annotator pairs, and no consistent tendency was observed.

\begin{table}[t]
\small
\centering
\begin{tabular}{cl|rr|rr}
\toprule
\multirow{2}{*}{Type} & \multirow{2}{*}{In/OOKB} 
& \multicolumn{2}{c|}{A vs. B} & \multicolumn{2}{c}{A vs. C} \\
&& \multicolumn{1}{c}{F1} & \multicolumn{1}{c|}{$\kappa$} & \multicolumn{1}{c}{F1} & \multicolumn{1}{c}{$\kappa$} \\
\midrule
\multirow{3}{*}{Exact}
& All & 0.924 & 0.919 & 0.830 & 0.817\\
& InKB & 0.930 & --    & 0.802 & --\\
& OOKB & 0.898 & --    & 0.917 & --\\
\midrule
Exact & All & 0.870 & 0.866 & 0.769 & 0.759 \\
\& & InKB & 0.881 & --    & 0.767 & -- \\
Related & OOKB & 0.800 & --    & 0.776 & -- \\
\bottomrule
\end{tabular}
\caption{Inter-annotator agreement for link annotation.} 
\label{tab:iaa_link}
\end{table}

\section{Detailed Experimental Settings} \label{sec:exp_detail}
\paragraph*{SPARQL Query} 
Figure~\ref{fig:sparql_query} shows the SPARQL query used in the preliminary experiment in \S\ref{sec:pre_eval} to retrieve the candidate entry set for each mention.

\begin{figure}[h]
\small
\begin{verbatim}
SELECT DISTINCT ?item ?itemLabel \
?itemDescription ?itemAltLabel ?jawiki WHERE {
  {?item rdfs:label "{text}"@ja .}
  UNION {?item skos:altLabel "{text}"@ja .}
  OPTIONAL {
    ?jawiki schema:about ?item .
    ?jawiki schema:inLanguage "ja" .
    FILTER (SUBSTR(str(?jawiki), 1, 25) = \
    "https://ja.wikipedia.org/")
  }
  SERVICE wikibase:label {
    bd:serviceParam wikibase:language "ja,en" .
  }
}
\end{verbatim}
\caption{SPARQL query used in the experiment.}
\label{fig:sparql_query}
\end{figure}

\paragraph*{Definition of Expected Recall} \label{sec:rec_def}
The expected value of Recall@$k$ for a system output was computed as follows.
Given a predicted entry set $E_p$ for mention $m$, where each entry $e \in E_p$ has a score $s$, entries are first taken in descending order of score to obtain the smallest subset $E_p^\prime \subset E_p$ such that $|E_p^\prime| \geq k$ (if $|E_p| < k$, then $E_p^\prime = E_p$).
If $|E_p^\prime| > k$, then $E_p^{\prime\prime}$ is obtained by randomly removing $|E_p^\prime|-k$ entries from among those entries in $E_p^\prime$ that have the lowest score, so that the resulting set has exactly $k$ entries (if $|E_p^\prime| \leq k$, then $E_p^{\prime\prime} = E_p^\prime$).
We then define $r_m$, the expected value of Recall@$k$ for mention $m$, as the probability that $E_p^{\prime\prime}$ contains the gold entry $e_g$.
The expected value of Recall@$k$ for the entire system output is obtained by averaging $r_m$ over all mentions $m \in M$, i.e., $\frac{1}{|M|} \sum_{m \in M} r_m$.

More specifically, for a mention $m$, when $e_g \in E_p$, $r_m$ is computed by the following equation (when $e_g \not\in E_p$, $r_m = 0$).
\begin{eqnarray*}
\small
\left\{
\begin{array}{ll}
1 & (\textrm{if}~|E_p(s \!\geq\! s_g)|\leq k)\\
\frac{k-|E_p(s > s_g)|}{|E_p(s = s_g)|} & (\textrm{if}~|E_p(s \!\geq\! s_g)|>k~\land~|E_p(s \!>\! s_g)|<k)\\
0 & (\textrm{otherwise})
\end{array}
\right.
\end{eqnarray*}
Here, $s_g$ denotes the score of the gold entry $e_g$, and $E_p(s \geq s_g)=\{e \in E_p \mid s \geq s_g\}$ denotes the set of entries in $E_p$ whose scores are greater than or equal to $s_g$.
The sets $E_p(s > s_g)$ and $E_p(s = s_g)$ are defined similarly.

\end{document}